\pdfoutput=1
\documentclass[11pt]{article}

\usepackage[final]{acl}

\usepackage{times}
\usepackage{latexsym}
\usepackage{natbib}

\usepackage[T1]{fontenc}
\usepackage[utf8]{inputenc}

\usepackage{microtype}

\usepackage{inconsolata}

\usepackage{graphicx}

\usepackage{amsmath}
\usepackage{enumitem}
\usepackage{booktabs}
\usepackage{amsthm}
\usepackage{balance}
\usepackage{soul}
\usepackage{url}
\usepackage[utf8]{inputenc}

\title{MIRA: Empowering One-Touch AI Services on Smartphones with MLLM-based Instruction Recommendation}

\author{
\textbf{Zhipeng Bian\textsuperscript{1,2}},~
\textbf{Jieming Zhu\textsuperscript{2}$\footnotemark[1]$},~
\textbf{Xuyang Xie\textsuperscript{2}},~
\textbf{Quanyu Dai\textsuperscript{2}},~
\textbf{Zhou Zhao\textsuperscript{3}}, \\
\textbf{Zhenhua Dong\textsuperscript{2}} \\
\textsuperscript{1}Shenzhen University, Shenzhen, China \quad \textsuperscript{2}Huawei Noah's Ark Lab, Shenzhen, China \\
\textsuperscript{3}Zhejiang University, Hangzhou, China \\
\texttt{bianzhipeng2022@email.szu.edu.cn} \quad \texttt{jiemingzhu@ieee.org}\\
\texttt{\{xiexuyang,daiquanyu,dongzhenhua\}@huawei.com} \quad \texttt{zhaozhou@zju.edu.cn} 
}

\begin{document}
\maketitle

\renewcommand{\thefootnote}{\fnsymbol{footnote}}
\footnotetext[1]{\ Corresponding Author.}

\begin{abstract}
The rapid advancement of generative AI technologies is driving the integration of diverse AI-powered services into smartphones, transforming how users interact with their devices. To simplify access to predefined AI services, this paper introduces MIRA, a pioneering framework for task instruction recommendation that enables intuitive one-touch AI tasking on smartphones. With MIRA, users can long-press on images or text objects to receive contextually relevant instruction recommendations for executing AI tasks. Our work introduces three key innovations: 1) A multimodal large language model (MLLM)-based recommendation pipeline with structured reasoning to extract key entities, infer user intent, and generate precise instructions; 2) A template-augmented reasoning mechanism that integrates high-level reasoning templates, enhancing task inference accuracy; 3) A prefix-tree-based constrained decoding strategy that restricts outputs to predefined instruction candidates, ensuring coherent and intent-aligned suggestions. Through evaluation using a real-world annotated datasets and a user study, MIRA has demonstrated substantial improvements in the accuracy of instruction recommendation. The encouraging results highlight MIRA's potential to revolutionize the way users engage with AI services on their smartphones, offering a more seamless and efficient experience.
\end{abstract}

\section{Introduction}
\label{introduction}
\begin{figure}
    \centering
    \includegraphics[width=\linewidth]{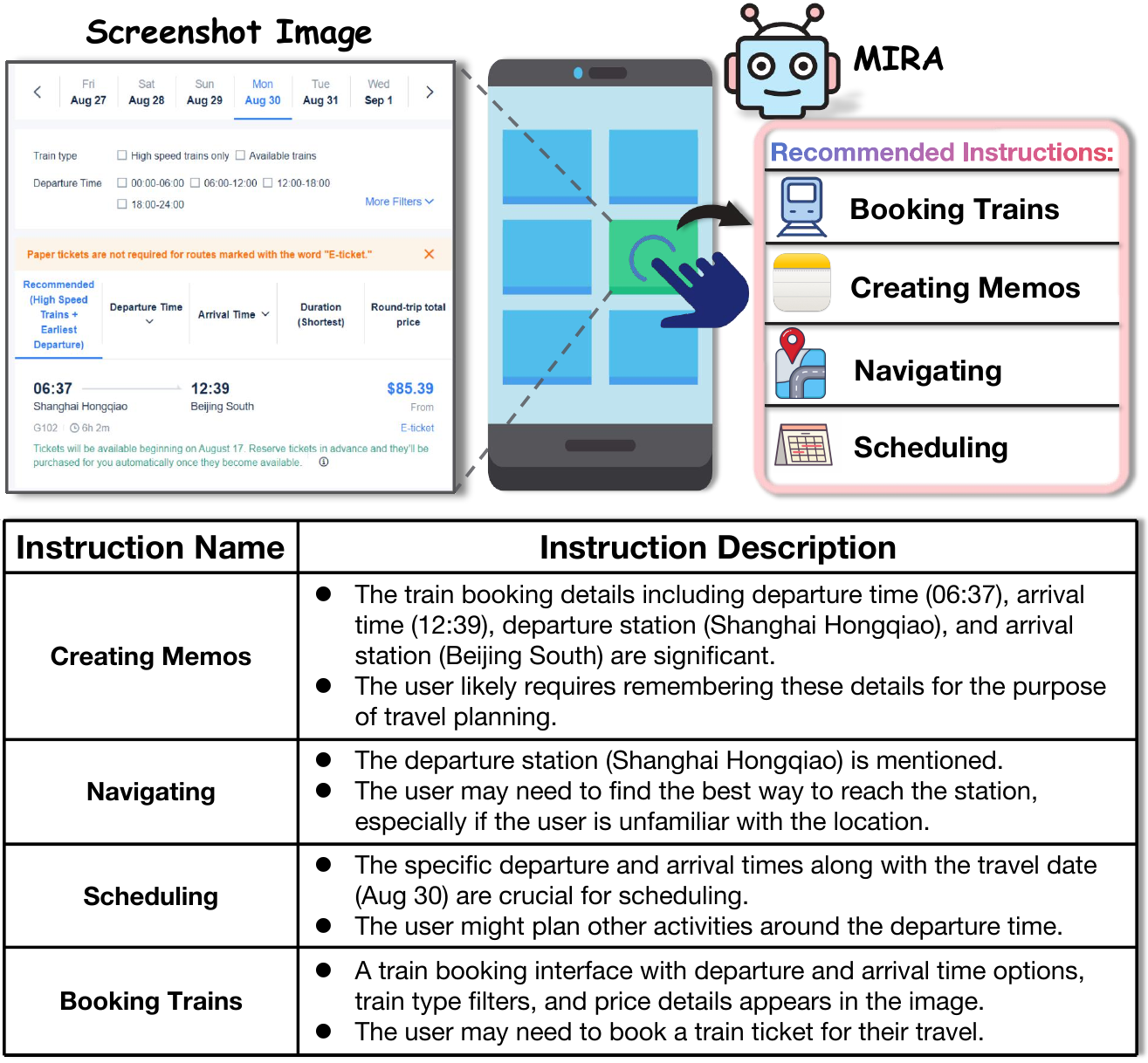}
    \caption{An illustration of one-touch AI services on smartphones.}
    \label{fig:intro}
\end{figure}

Generative AI technologies, such as large language models (LLMs)~\cite{LLMSurvey}, diffusion models~\cite{DiffusionSurvey}, and AI agents~\cite{AIAgent}, are revolutionizing the capabilities of AI smartphones~\cite{AI_Smartphone, AI_Smartphone2}, ushering in a new era of intelligent mobile devices that offer unparalleled levels of personalization and interaction. The integration of LLMs powers sophisticated virtual assistants capable of engaging in more natural, contextually rich conversations, delivering detailed and relevant information to meet specific user needs. These advancements also enable smartphones to generate high-quality images, videos, text, and music on demand, providing users with unprecedented creative freedom. Furthermore, it empowers a wide range of AI services, including real-time language translation, advanced image captioning, visual question answering, customized text summaries, and personalized recommendations, setting a new standard for smartphone functionality. As generative AI continues to evolve, it promises seamless integration into various aspects of smartphone use, transforming mobile devices into intelligent AI agents that adeptly serve daily needs.

Given the rich AI capabilities on smartphones, there is significant potential for seamless and effortless AI services. Currently, most smartphones rely on conversational AI assistants (e.g., Siri) to process user requests via text or voice commands. While effective for interaction, this approach has limitations in handling routine daily tasks. Users often need detailed, multi-step instructions to complete AI tasks. For example, processing a screenshot of a train booking involves several steps: performing text recognition (i.e., OCR), extracting structured information, adding the event to a calendar, and setting up a reminder. This process is time-consuming and cumbersome. Additionally, repeatedly executing these instructions for daily repetitive tasks wastes valuable time and effort.

To address these challenges and promote seamless access to AI services, we propose MIRA, a Multimodal Instruction Recommendation Agent that enables one-touch AI task execution on smartphones. Users can long-press target objects such as images, messages, or documents, triggering pre-defined, contextually relevant task instruction recommendations for accessing AI services. For example, as shown in Figure \ref{fig:intro}, when handling a screenshot of a train booking, a user can simply long-press the image to instantly receive recommendations for actions like booking trains, creating memos, scheduling calendar events, and navigating to station. In this paper, we define AI task instructions as detailed prompts and execution steps to operate on trigger objects and complete a specific task. For instance, a navigation task might involve recognizing a specific location from an image and subsequently invoking a map navigation API to guide the user. By encapsulating complex processes into single, intuitive actions, MIRA allows users to quickly and effortlessly access AI services, simplifying task completion and maximizing convenience.

This represents an emerging application scenario in the new era of AI smartphones. To our knowledge, it is the first effort to address instruction recommendations for AI services on smartphones. With the rise of generative AI, functionalities such as translation, summarization, navigation, event scheduling, calling, memo creation, image editing, image description, calorie calculation, and cooking inquiries are now available. Each service typically involves a complex pipeline of prompts, fine-tuned models (e.g., LoRAs), and API calls. These services can be added by smartphone providers or registered by third-party partners. MIRA's main goal is to provide contextually relevant recommendations from a wide range of AI services when users long-press a specific image or text object (i.e., triggers). While supporting various trigger types is ideal, we focus on text and image triggers in this initial effort.

Unlike traditional recommender systems that focus on user behavior sequences, our instruction recommendation task emphasizes on multimodal trigger inputs. The challenge lies in understanding the content of these triggers and extracting key information to infer user intent and generate precise instructions. For example, given an image of a bank card, the system should recognize tasks like transferring funds or creating memos related to banking.

This paper introduces MIRA, a multimodal large language model (MLLM)-based recommendation agent for understanding user context and recommending task instructions. While MLLMs excel in image recognition and text understanding~\cite{liu2024ocrbench},aligning trigger content with relevant instructions is challenging. We make three key contributions: 1) Introducing structured reasoning to extract entities, infer user intent, and generate precise instructions; 2) Developing a template-augmented reasoning mechanism to improve task inference accuracy; 3) Implementing prefix-tree-based constrained decoding to ensure coherence and intent alignment. We evaluate MIRA using real-world datasets and a user study, showing significant improvements in instruction recommendation accuracy.

\begin{figure*}
    \centering
    \begin{minipage}[b]{\linewidth}
        \centering
        \centerline{\includegraphics[width=\linewidth]{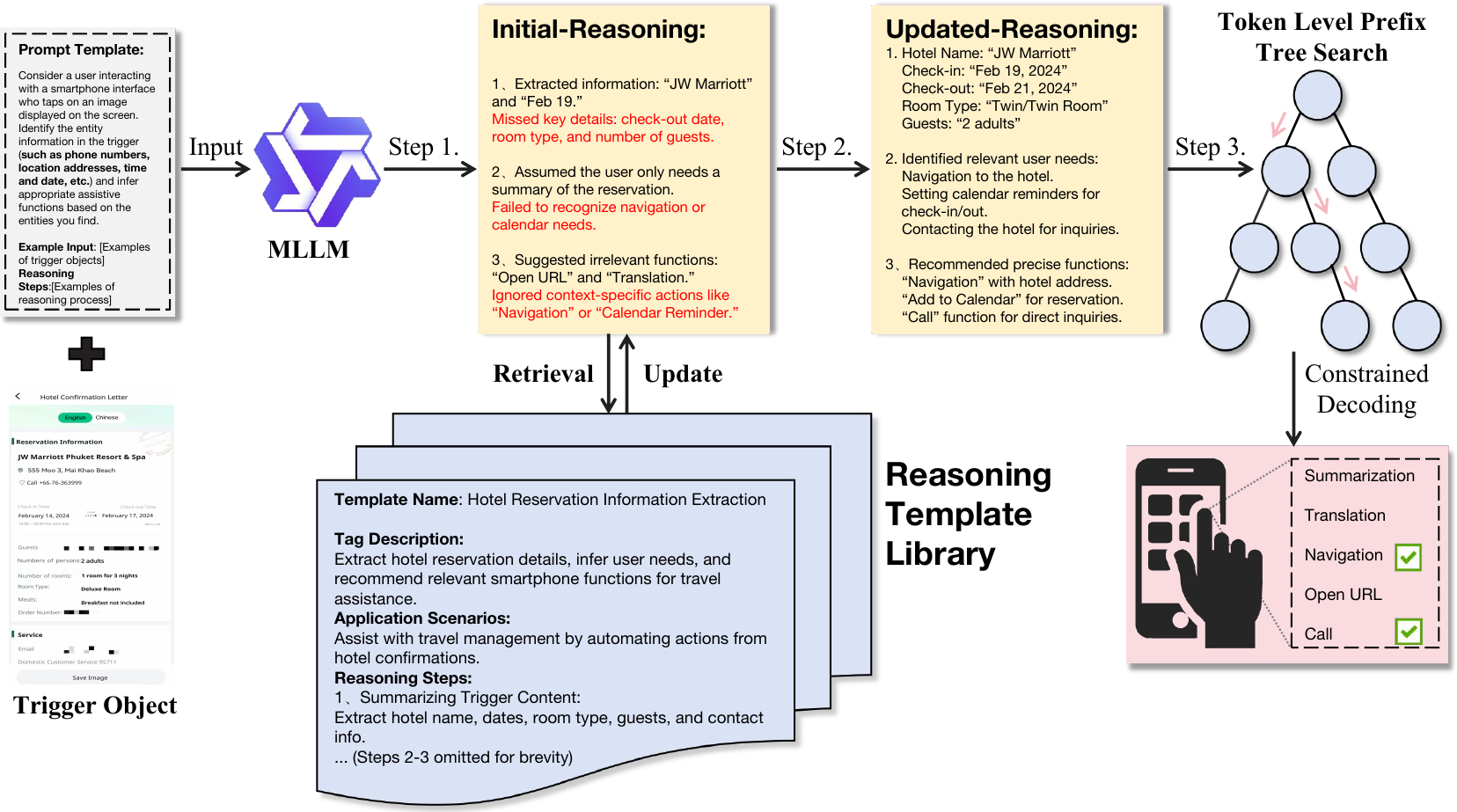}}
    \end{minipage}
    \caption{Overview of MIRA. A prompt template and trigger object extract structured information through initial reasoning, refine reasoning steps through template retrieval and updates, and apply constrained decoding during inference to recommend predefined instructions.}
    \label{fig:method}
\end{figure*}
\section{Related Work}
\subsection{Multimodal Large Language Model Reasoning}
Recent advances in MLLM have highlighted their impressive visual reasoning capabilities. Studies have explored plan-based Chain-of-Thought (CoT) prompting~\cite{shao2024visual, mitra2024compositional}, which guides models through intermediate reasoning steps for more accurate results. LLaVA-CoT~\cite{xu2024llava} introduces a Vision-Language Model (VLM) designed for structured reasoning, achieving notable success in visual tasks. Building on this, LlamaV-o1~\cite{thawakar2025llamav} uses multi-stage curriculum learning to progressively improve problem-solving skills. CoMCTS~\cite{yao2024mulberry} combines collective learning with tree search to optimize reasoning pathways. However, these methods either require lengthy tree search algorithms or rely on process reward models to guide reasoning, making them inefficient. As a result, an efficient and effective approach for complex reasoning tasks in MLLMs is still lacking.

\subsection{MLLMs for Recommendation}
Recent studies have explored integrating MLLMs into multimodal recommendation systems~\cite{MMRecSurvey}, leveraging their ability to process diverse data modalities. Frameworks like VIP5~\cite{geng2023vip5} align visual, textual, and personalization cues to enhance performance with personalized prompts and efficient training. MLLM-MSR~\cite{ye2024harnessing} captures dynamic user preferences by summarizing multimodal inputs, while TMF~\cite{ma2024triple} improves multi-behavior recommendations by incorporating graph data. Recently, DeepMP~\cite{DeepMP} unifies multimodal recommendation and generation within a single MLLM model. These advancements underscore the potential of MLLMs to refine recommendations by analyzing user preferences across modalities. However, ensuring precise alignment between multimodal triggers and actionable AI services remains an open challenge.

\section{Methodology}
\label{method}
We present MIRA, a novel framework designed to enhance instruction recommendation tasks. As illustrated in Figure~\ref{fig:method}, MIRA comprises three key components: structured chain-of-thought reasoning, template-augmented structured reasoning, and prefix-tree-based constrained decoding.

\subsection{Enhancing MLLMs with Structured Reasoning}
\begin{figure}[t]
    \includegraphics[width=\linewidth]{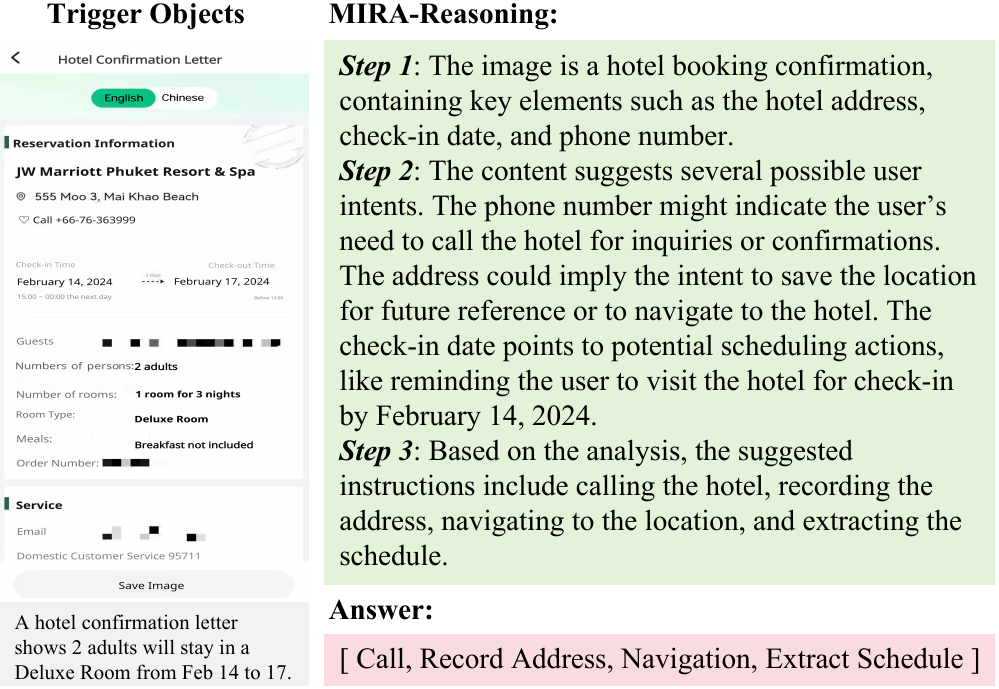}
    \caption{A Sample of the Reasoning-Dataset: Left - Image or Text Trigger Objects, Right - MIRA-Reasoning Process and Final Answers.} \label{fig:data}
\end{figure}
Multimodal large language models (MLLMs) excel in tasks like OCR, object detection, and image captioning but struggle with complex reasoning tasks involving implicit constraints and object relationships. In our task, MLLMs find it difficult to infer user intent from trigger objects and recommend instructions accurately. Inspired by OpenAI's O1~\cite{jaech2024openai}, DeepSeek-R1~\cite{guo2025deepseek}, and Qwen-QwQ~\cite{yang2024qwen2}, we enhance MLLMs with human-like reasoning, enabling thoughtful processing of trigger objects and precise recommendations. By leveraging zero-shot Chain-of-Thought (CoT) prompting in models like GPT-4V and Qwen2.5VL-Max, we incorporate structured reasoning into the trigger-instruction dataset, improving MLLMs' ability to match trigger content with instructions.

To guide the model, We designed a three-step reasoning trajectory to guide the model in processing trigger objects. First, in entity recognition and summarization, MLLMs extract key entities (e.g., phone numbers, addresses, dates) from text and images, organizing them into structured themes. Next, in the contextual relevance analysis, the model links these summaries to user intent, connecting entities like dates or locations to actions such as saving or navigating. Finally, in the instruction generation step, the model synthesizes the reasoning into context-aware, user-focused recommendations. Formally, for a single sample $S_i$ in the dataset, which consists of the trigger object $q_i$ and the ground truth instruction $a_i$, we provide $a_i$ directly to the MLLM, ensuring both consistency and precision. Given a rich prompt template with in-context examples $p^e_i$, MLLM constructs the reasoning steps $r_i$ based on the provided correct answer. The input and output format for the MLLM is as follows:
\begin{equation}
\label{eq:data_construction}
\begin{aligned}
  r_i = MLLM(p^e_i, q_i, a_i).
\end{aligned}
\end{equation}
Equation~\ref{eq:data_construction} represents the generation of high-quality reasoning traces under teacher forcing with gold answers, used to bootstrap the initial training dataset. After constructing the reasoning dataset as shown in Figure~\ref{fig:data}, we perform supervised fine-tuning (SFT) on MLLMs. During training, the model is provided only with the prompt  $p_i$ (without in-context examples) and the trigger object $q_i$, generating predicted reasoning steps  $\hat{r}_i$ and predicted answers  $\hat{a}_i$:
\begin{equation}
\label{eq:training}
\begin{aligned}
  \hat{r}_i\;, \hat{a}_i = MLLM(p_i, q_i).
\end{aligned}
\end{equation}
Equation~\ref{eq:training} shows how the trained MLLM learns to independently produce reasoning and instructions given only trigger context, improving autonomy. This approach equips the MLLM with reasoning capabilities for complex tasks while eliminating the need for large-scale models and intricate prompt engineering. Specifically, we introduce two special tokens, <$REASONING$> and <$/REASONING$>, marking the start and end of the reasoning process, thereby enabling autonomous reasoning.

\subsection{Template-Augmented Structured Reasoning}
\label{sec:template}
After fine-tuning on a reasoning dataset, MLLMs are capable of structuring reasoning to analyze complex content and relationships of trigger objects, enabling accurate instruction recommendations. However, the accuracy of this reasoning is challenged by inherent randomness and hallucination tendencies, with no explicit supervision to ensure the correctness of the reasoning steps~\cite{zhang2025does}.

To address this, we propose the Reasoning Template Library. This library uses high-level, solution-oriented templates for structured reasoning, reducing inaccuracies and inconsistencies. Built using closed-source MLLMs' summarization capabilities, it distills common problem-solving patterns from the dataset. By identifying recurring strategies, we developed robust templates that ensure efficient and precise instruction recommendations for diverse trigger objects.

As shown in the lower center of Figure~\ref{fig:method}, each template includes four key components in a structured metadata format: \textbf{Template Name} (e.g., "Hotel Reservation Information Extraction"), \textbf{Tag Description} with keywords for easy search (e.g., "Travel," "Reservation," "Hotel"), a brief summary of \textbf{Application Scenarios}, and \textbf{Reasoning Steps} outlining reasoning steps (e.g., "Extract hotel name," "Identify check-in date," "Recommend calendar reminder"). This metadata enables efficient retrieval, ensuring quick, accurate searches based on keywords or problem characteristics for relevant templates.

As shown in Figure~\ref{fig:method}, after building the reasoning template library, the next step is integrating these templates with the MLLM to enhance its reasoning. The process begins when the trigger object (e.g., a hotel reservation confirmation) is provided to the MLLM, which generates initial reasoning outlining task steps. However, in complex scenarios, this reasoning may be incomplete. To address this, we use vector-based retrieval to find the most relevant template by calculating the similarity between the initial reasoning vector and each template's vector, formalized as:
\begin{equation}
\label{eq:similarity}
\begin{aligned}
    j = \text{argmax}_i(\text{Sim}(f(\hat{r}),\{f(D_{T_i})\}_{i=1}^N)),\\ \text{where}\quad \text{Sim}(f(\hat{r}),\{f(D_{T_i})\}_{i=0}^n) >= \delta.
\end{aligned}
\end{equation}
where $f(\hat{r})$ represents the embedding of the initial reasoning, and $\{f(D_{T_i})\}_{i=1}^N$ represents the embeddings of the templates in the library. $Sim(\cdot,\cdot)$ is the similarity function, which measures how closely the reasoning steps of each template align with the task at hand. We set a threshold $\delta$ (recommended range: 0.5–0.7) to ensure the selected template is suitable for the given trigger object. The most relevant template $T_j$ is selected, and its reasoning steps are used to update the initial reasoning.

To ensure adaptability in dynamic smartphone usage scenarios, our template library supports continual evolution. During inference, when a trigger object results in low similarity to all existing templates (i.e., no suitable template passes the similarity threshold $\delta$), we log the reasoning trace generated by the MLLM as a candidate for future template distillation. These reasoning traces are periodically clustered based on semantic similarity, and representative examples are selected and summarized by Qwen2.5VL-Max into new candidate templates. Before adding any newly distilled template $D_{T_\text{new}}$ to the library, we compute its embedding $f(D_{T_\text{new}})$ and compare it against the existing templates $\{f(D_{T_i})\}_{i=1}^{n}$. A new template is added only if the maximum similarity is below a threshold $\delta$, ensuring informativeness and non-redundancy:
\begin{equation}
\label{eq:update_rule}
\begin{aligned}
    \max \left( \text{Sim}(f(D_{T_\text{new}}), \{f(D_{T_i})\}_{i=1}^{n}) \right) < \delta.
\end{aligned}
\end{equation}
Here, $\text{Sim}(\cdot,\cdot)$ denotes the cosine similarity between two embeddings, and $\delta$ is typically set to 0.5 to balance coverage and redundancy. This condition helps prevent duplicate entries, ensuring that only novel and informative templates are added. As a result, the template library can continuously evolve over time, capturing new reasoning strategies and accommodating rare edge-case scenarios encountered during real-world deployment.

The final step is to instantiate the reasoning by inputting the retrieved template and trigger object into the MLLM to generate the updated reasoning steps. This can be represented as:
\begin{equation}
\label{eq:update}
\begin{aligned}
    \hat{r}_{updated} \leftarrow MLLM(T_{j}, q_i).
\end{aligned}
\end{equation}
where $\hat{r}_{updated}$ represents the updated reasoning steps. Equation~\ref{eq:update} represents the final reasoning refinement, injecting template guidance into the inference trajectory.

This process refines reasoning to better align with task requirements. For example, with a hotel reservation trigger object, the initial MLLM reasoning might only extract the hotel name and check-in date. By retrieving a relevant template, the reasoning is enriched with details like check-out date, room type, number of guests, and actions such as setting a reminder or providing navigation. This template-driven approach improves accuracy, reduces computational demands, and enables easier deployment without additional training.

%Leveraging the reasoning dataset shown in Figure~\ref{fig:data}, we perform supervised fine-tuning (SFT) on MLLMs. During training, the model receives only the prompt $p_i$ (without in-context examples) and the trigger object $q_i$, outputting the reasoning steps $\hat{r}_i$ and predicted answers $\hat{a}_i$:

%This method maintains CoT reasoning flexibility while removing the need for large models. During fine-tuning, the focus is on improving reasoning capabilities without complex prompting. Special tokens, <$REASONING$> and <$/REASONING$>, mark the reasoning steps, enabling autonomous reasoning. This approach boosts performance in instruction recommendation without external prompt engineering.

\begin{figure}[t]
    \includegraphics[width=\linewidth]{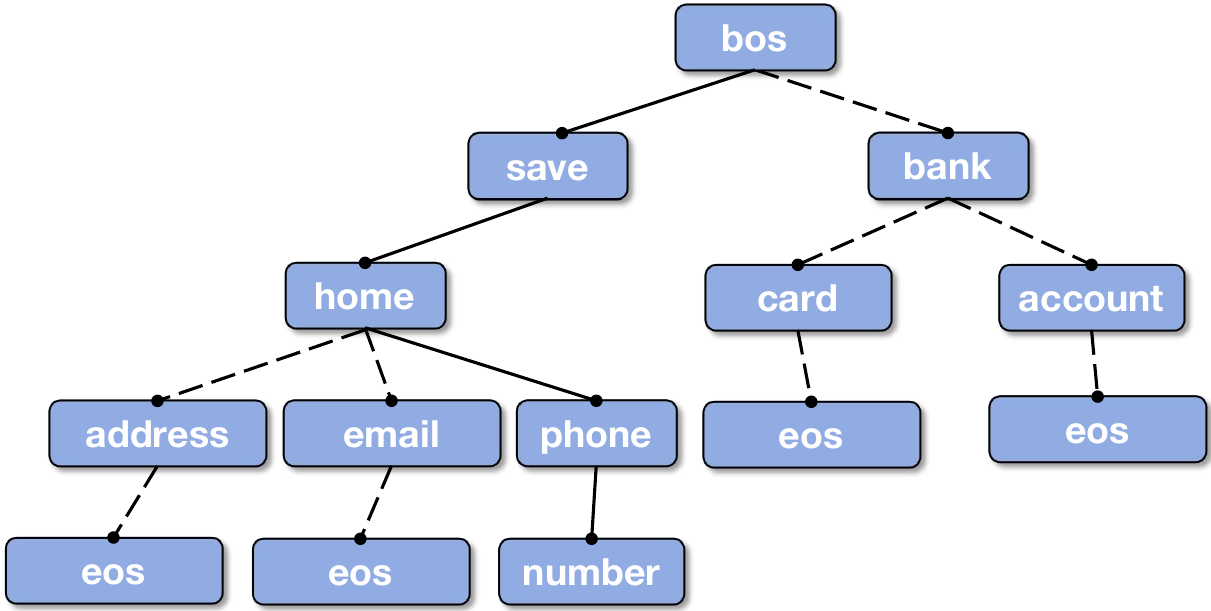}
    \caption{The Illustration of Prefix Tree Searching.} \label{fig:prefix_tree}
\end{figure}

\subsection{Prefix-Tree-Based Constrained Decoding}
To prevent the model from generating irrelevant instructions during inference, we implement constrained decoding with a prefix tree built from the MLLM's tokenizer and candidate instructions. After the end token (<$/REASONING$>) of the reasoning process, the model switches to the prefix tree, masking logits for invalid tokens and ensuring only valid sequences are generated, as shown in Figure~\ref{fig:prefix_tree}. To build the prefix tree, we tokenize all valid instruction sequences using the MLLM’s tokenizer and recursively construct trie nodes, marking valid transitions. During inference, the decoder filters logits using the current tree node’s valid token set, ensuring efficient decoding. The tree is rebuilt dynamically when the instruction library updates and supports O(L) time decoding per token, where L is the sequence length. For example, selecting “save” leads to options like “home,” “address,” “email,” and “phone.” It then selects “phone,” followed by “number,” forming the instruction “save phone number.” This approach eliminates post-processing and reduces MLLMs' hallucination, ensuring precise outputs.

\section{Experiments}
\subsection{Datasets and Template Library Construction}
To train and validate MIRA, we built a dataset from 1,000 smartphone users, representing diverse demographics and usage patterns. Each sample was annotated by at least three users, ensuring robust inter-annotator agreement ($\kappa$ = 0.85). The dataset includes 4,952 training pairs and 956 testing pairs, providing a reliable basis for evaluating MIRA’s performance in real-world scenarios.
\begin{table}[h]
\begin{minipage}[b]{\linewidth}
\centering
\resizebox{\linewidth}{!}{
\begin{tabular}{llrrrrrrrrrrrrr}
\toprule
Model & Method  & Recall & Precision & F1-score & HR@1 & HR@3\\
\midrule
InternVL2.5-2B  & Zero-shot & 0.2904  & 0.3042 & 0.2971  & 0.3829 & 0.4012\\
 & Vanilla-SFT  & 0.4115  & 0.4201  & 0.4158  & 0.4942 & 0.5052\\
 & MIRA  & 0.7164  & 0.7382  & 0.7271  & 0.8051 & 0.8351\\
\midrule
Qwen2.5VL-2B & Zero-shot & 0.3043  & 0.3207  & 0.3122  & 0.3941 & 0.4223\\
 & Vanilla-SFT  & 0.4964  & 0.4882  & 0.4923  & 0.5051 & 0.5321\\
 & MIRA  & 0.7489 & 0.7397 & 0.7443  & 0.8151 & 0.8451\\
\midrule
InternVL2.5-8B & Zero-shot & 0.3145  & 0.3319  & 0.3230  & 0.4512 & 0.4783\\
 & Vanilla-SFT  & 0.5254 & 0.5827  & 0.5526  & 0.5963 & 0.6128\\
 & MIRA  & {\ul {0.9283}}  & {\ul {0.9154}}  & \textbf{0.9218}  & {\ul {0.9354}} & {\ul {0.9516}}\\
\midrule
Qwen2.5VL-7B & Zero-shot & 0.3294  & 0.3424  & 0.3358  & 0.4589 & 0.4924\\
 & Vanilla-SFT  & 0.5678  & 0.5731  & 0.5704  & 0.6012 & 0.6841\\
 & MIRA  & \textbf{0.9286}  & \textbf{0.9239}  & {\ul {0.9121}}  & \textbf{0.9542} & \textbf{0.9629}\\
\midrule
\end{tabular}
}
\end{minipage}
\caption{Quantitative Comparisons Between MIRA and Baseline Methods: The best results are in \textbf{bold}, and the second-best results are {\ul {underlined}}. All metrics indicate better performance with higher values.}
\label{tab:performance}
\end{table}
As detailed in Section 3.2, we use Qwen2.5VL-Max to extract high-level insights from the training data, which are then used to construct a structured thought template library containing approximately 80 templates. We utilize \textbf{jina-embeddings-v3}\footnote{https://huggingface.co/jinaai/jina-embeddings-v3.} for template retrieval. While we leverage a closed-source model for high-level insight extraction during template construction, the reasoning patterns distilled from these summaries are model-agnostic and serve as generalizable abstractions for diverse scenarios. In future work, we plan to incorporate open-source models and crowdsourced annotations to enhance cross-model generality and robustness.

\subsection{Baselines and Metrics}
In our experiments, we did not compare with MLLM4Rec or LLM4Rec methods, such as MLLM-MSR~\cite{ye2024harnessing}, Rec-GPT4V~\cite{liu2024rec}, LLMRank~\cite{hou2024large}, and NoteLLM-2~\cite{zhang2024notellm}, as they focus on sequence recommendation tasks requiring user behavior data. These methods target different tasks than our instruction recommendation framework. Instead, we compared MIRA with two baseline methods: zero-shot prompting with in-context learning~\cite{dong2022survey} and supervised fine-tuning on the original dataset. These methods are more aligned with our task and serve as a relevant benchmark for evaluating MIRA's performance. Experiments were conducted on four MLLMs: InternVL2.5~\cite{chen2024expanding} (2B and 8B) and Qwen2.5VL~\cite{bai2023qwen} (2B and 7B). We evaluated MIRA using four standard recommendation system metrics: recall, precision, F1-score, and hit rate. All experiments were performed on two GPUs with 32GB of memory.

\subsection{Experimental Results}
\subsubsection{\textbf{Comparison with baselines.}}
Table~\ref{tab:performance} presents the model performance, where MIRA consistently outperforms baseline methods across models of varying sizes: InternVL2.5 (2B and 8B) and Qwen2.5VL (2B and 7B). Notably, MIRA achieves substantial improvements across all metrics. For instance, on Qwen2.5VL-7B, MIRA reaches a macro F1-score of 0.9121 and HR@3 of 0.9629, significantly surpassing Vanilla-SFT. These improvements stem from MIRA's modular architecture, enhancing reasoning and ensuring accurate recommendations. Furthermore, MIRA's superior performance on smaller models like InternVL2.5-2B and Qwen2.5VL-2B underscores its efficiency, making it well-suited for real-world deployment with constrained resources.
\begin{table}[h]
\begin{minipage}[b]{\linewidth}
\centering
\resizebox{\linewidth}{!}{
\begin{tabular}{lccc}
\toprule
Model & initial reasoning  & with Template \\
\midrule
InternVL2.5-2B  & 0.6041 & 0.7271 \textbf{($\uparrow$ 20.4\%)}\\

Qwen2.5VL-2B & 0.6428 & 0.7443 \textbf{($\uparrow$ 15.8\%)}\\

InternVL2.5-8B & 0.7451 & 0.9218 \textbf{($\uparrow$ 23.7\%)}\\

Qwen2.5VL-7B & 0.7348 & 0.9121 \textbf{($\uparrow$ 24.1\%)}\\
\midrule
\end{tabular}
}
\end{minipage}
\caption{Ablation study on the impact of template-augmented reasoning on instruction recommendation performance.}
\label{tab:a1}
\end{table}
\subsubsection{\textbf{Depth Analysis.}}
An ablation study was conducted to evaluate the impact of the template-augmented structured reasoning method. The study compares instruction recommendation performance using initial reasoning versus template-enhanced reasoning, measured by the F1 score, as shown in Table~\ref{tab:a1}. The results demonstrate a significant improvement with template-enhanced reasoning. InternVL2.5-2B saw a 20.4\% increase, Qwen2.5VL-2B improved by 15.8\%, reaching 0.7443, while larger models showed even greater gains: InternVL2.5-8B improved by 23.7\%, and Qwen2.5VL-7B by 24.1\%, reaching 0.9121. Template retrieval mitigates hallucination issues common in unsupervised reasoning, significantly boosting recommendation accuracy.

We further evaluated MIRA (Qwen2.5VL 7B) against two state-of-the-art multimodal large language models—Qwen2.5VL-Max and GPT-4V—both commonly deployed via API in industrial applications. All models were tested using the same trigger objects and full templates under a zero-shot Chain-of-Thought (CoT) prompting setup (“Let’s think step by step”)~\cite{kojima2022large}. The evaluation considered four key metrics: F1-score, average token length, inference time, and model size. As shown in Table~\ref{tab:a2}, MIRA achieved the highest F1-score of 0.9121, outperforming both GPT-4V (0.879) and Qwen2.5VL-Max (0.861). It also demonstrated superior token efficiency, requiring only 116 tokens on average—far fewer than GPT-4V (817) and Qwen2.5VL-Max (807). Despite having just 7 billion parameters, MIRA completed inference in 11.2 seconds, faster than GPT-4V (11.3s) and comparable to Qwen2.5VL-Max (10.7s). These results highlight MIRA’s strong balance between accuracy and efficiency. Its compact architecture enables faster and lighter inference without compromising performance, making it highly suitable for deployment in resource-constrained environments such as smartphones and edge devices—where both responsiveness and computational cost are critical.
\begin{table}[h]
\begin{minipage}[b]{\linewidth}
\centering
\resizebox{\linewidth}{!}{
\begin{tabular}{lccccc}
\toprule
Model & F1-score  & Token Length &Inference Time & Model Parameters\\
\midrule
GPT-4V  & 0.879 & 817 &11.3s  & >500B\\

Qwen2.5VL-Max &0.861  & 807 &\textbf{10.7s}  & >500B\\

MIRA  & \textbf{0.9121} & \textbf{116} &11.2s  & \textbf{7B}\\
\midrule
\end{tabular}
}
\end{minipage}
\caption{Analysis of MIRA compared to Qwen2.5VL-Max and GPT-4V on key industrial metrics: The best results are in \textbf{bold}.}
\label{tab:a2}
\end{table}

To further investigate the robustness of the template matching process, we conducted a sensitivity analysis on the similarity threshold $\delta$ used in Equation~\ref{eq:similarity}. We evaluated instruction recommendation performance using the F1-score as the primary metric, varying $\delta$ across multiple settings. As shown in Table~\ref{tab:delta}, $\delta = 0.6$ consistently yields the highest F1-score across different MLLMs. Lower thresholds (e.g., $\delta = 0.4$) tend to retrieve overly generic templates, leading to irrelevant or misaligned reasoning steps. In contrast, higher thresholds (e.g., $\delta = 0.8$) significantly reduce the number of matched templates, resulting in degraded performance due to limited reasoning support. These results highlight the importance of properly tuning $\delta$ to balance retrieval coverage and reasoning precision.

\begin{table}[h]
\begin{minipage}[b]{\linewidth}
\centering
\resizebox{\linewidth}{!}{
\begin{tabular}{lcccc}
\toprule
Model & $\delta = 0.4$ & $\delta = 0.5$ & $\delta = 0.6$ & $\delta = 0.8$ \\
\midrule
InternVL2.5-2B & 0.6892 & 0.7145 & \textbf{0.7271} & 0.7008 \\
Qwen2.5VL-2B & 0.7014 & 0.7312 & \textbf{0.7443} & 0.7221 \\
InternVL2.5-8B & 0.8893 & 0.9122 & \textbf{0.9218} & 0.9051 \\
Qwen2.5VL-7B & 0.8945 & 0.9012 & \textbf{0.9121} & 0.8958 \\
\bottomrule
\end{tabular}
}
\end{minipage}
\caption{Sensitivity analysis of the similarity threshold $\delta$ for template retrieval. The best results are in \textbf{bold}.}
\label{tab:delta}
\end{table}

\subsubsection{\textbf{Failure Case Analysis.}}
We examined 100 incorrect predictions from MIRA to understand common failure patterns. Three major types emerged: (1) Entity Omission: MLLMs occasionally ignore subtle entities like timestamps in footnotes; (2) Template Misalignment: Vector retrieval retrieves a loosely relevant template, leading to incorrect reasoning paths; (3) Ambiguity in Triggers: When triggers contain overlapping intent signals (e.g., calendar + contacts), MIRA may prioritize one over the other. Future improvements will incorporate multi-template aggregation and confidence-based filtering.

\subsubsection{\textbf{User study.}}
We invited 100 participants to evaluate 500 trigger objects, each with 1 to 3 instruction recommendations generated by two MIRA versions based on Qwen2.5VL-7B and InternVL2.5-7B. Participants selected recommendations that aligned with their expectations. The evaluation metric was the validity ratio, defined as the proportion of selected recommendations meeting participants' expectations out of the total provided. Our method achieved validity ratios of 93\% and 95\% for the two versions, respectively, demonstrating its real-world effectiveness.

\section{Conclusion}
We proposed MIRA, a framework leveraging MLLMs for instruction recommendations on smartphones. By enabling users to obtain task suggestions through a simple long-press on images or text, MIRA streamlines AI task execution, reducing cognitive load and enhancing user interaction efficiency. Key innovations include structured reasoning, template-augmented reasoning, and prefix-tree-based constrained decoding, which enhance recommendation accuracy and consistency. Experiments and user studies show that MIRA outperforms existing methods, offering efficient resource use and positioning it as an ideal solution for AI service integration on mobile devices.

\section*{Limitations}
While MIRA offers substantial improvements in multimodal instruction recommendation, several limitations remain that point to promising directions for future research.

First, \textbf{the current trigger modality coverage is limited}. MIRA primarily supports text and image inputs, which restricts its applicability in more diverse smartphone contexts involving audio, video, or sensor data. To expand its generality, we plan to explore multimodal extensions that incorporate audio transcriptions (e.g., voicemail), video scene understanding (e.g., meeting highlights), and sensor signals (e.g., location or step count), enabling richer and more adaptive instruction recommendations.

Second, \textbf{the reliance on a predefined template library may constrain adaptability}. While the template-augmented structured reasoning mechanism significantly enhances accuracy, its performance may degrade on previously unseen or long-tail tasks. Although we adopt a dynamic update mechanism to evolve the template library (see Section~~\ref{sec:template}), the approach still depends on effective template coverage and accurate retrieval. Additional improvements such as multi-template aggregation or fallback strategies may be needed to enhance generalization.

Third, \textbf{real-world deployment raises issues of robustness, scalability, and privacy}. Despite reducing hallucination through constrained decoding and template guidance, MIRA may still encounter reasoning errors in highly complex or ambiguous triggers. Moreover, its effectiveness hinges on high-quality and diverse training data, especially for capturing rare user intents or edge cases. Lastly, since MIRA operates on potentially sensitive content like images, documents, or messages, future deployments must ensure privacy through techniques such as on-device inference, secure model serving, and data anonymization. We also aim to explore differential privacy to further mitigate risk.

Overall, these limitations provide a roadmap for extending MIRA into a more flexible, reliable, and privacy-conscious framework in future work. 

\section*{Acknowledgments}
We thank MindSpore (\url{http://mindspore.cn}) for the partial support of this work, which is a new deep learning computing framework.

%\balance
\bibliography{acl_latex}
% \appendix

\end{document}